\documentclass[sigconf,screen,prologue,table]{acmart}

\acmSubmissionID{4}

\usepackage{multirow}    
\usepackage{graphicx}    
\usepackage{rotating}    
\usepackage{booktabs}    
\usepackage{amsmath}
\usepackage{tabularx}
\usepackage{booktabs}
\usepackage{amsmath}
\usepackage{caption}
\usepackage{subcaption}
\acmConference[ACM MM]{Make sure to enter the correct conference title from your rights confirmation email}{2025}{Dublin, Ireland}
\renewcommand{\shortauthors}{Hanzhe}
\author{Hanzhe Liang} 
\affiliation{
  \institution{College of Computer Science and Software Engineering, Shenzhen University} 
  \institution{Shenzhen Audencia Financial Technology Institute}
  \city{Shenzhen}
  \country{China}
}
\email{2023362051@email.szu.edu.cn}

\author{Jie Zhang}
\affiliation{
  \institution{Faculty of Applied Sciences, Macao Polytechnic University}
  \city{Macao}
  \country{China}
}
\email{jpeter.zhang@mpu.edu.mo}

\author{Tao Dai}
\affiliation{
  \institution{College of Computer Science and Software Engineering, Shenzhen University}
  \city{Shenzhen}
  \country{China}
}
\email{daitao@szu.edu.cn}

\author{Linlin Shen}
\affiliation{
  \institution{School of Artificial Intelligence, Shenzhen University}
  \institution{Guangdong Provincial Key Laboratory of Intelligent Information Processing}
  \city{Shenzhen}
  \country{China}
}
\email{llshen@szu.edu.cn}

\author{Jinbao Wang}
\affiliation{
  \institution{School of Artificial Intelligence, Shenzhen University}
  \institution{Guangdong Provincial Key Laboratory of Intelligent Information Processing}
  \city{Shenzhen}
  \country{China}
}
\email{wangjb@szu.edu.cn}

\author{Can Gao}
\affiliation{
  \institution{College of Computer Science and Software Engineering, Shenzhen University}
  \institution{Guangdong Provincial Key Laboratory of Intelligent Information Processing}
  \city{Shenzhen}
  \country{China}
}
\email{davidgao@szu.edu.cn}
\thanks{Corresponding author: Can Gao. Email: davidgao@szu.edu.cn}
\copyrightyear{2025}
\acmYear{2025}
\setcopyright{acmlicensed}\acmConference[MM '25]{Proceedings of the 33rd ACM International Conference on Multimedia}{October 27--31, 2025}{Dublin, Ireland}
\acmBooktitle{Proceedings of the 33rd ACM International Conference on Multimedia (MM '25), October 27--31, 2025, Dublin, Ireland}
\acmDOI{10.1145/3746027.3754492}
\acmISBN{979-8-4007-2035-2/2025/10}
\begin{document}
\title{Taming Anomalies with Down-Up Sampling Networks: Group Center Preserving Reconstruction for 3D Anomaly Detection}

\begin{abstract}
Reconstruction-based methods have demonstrated very promising results for 3D anomaly detection. However, these methods face great challenges in handling high-precision point clouds due to the large scale and complex structure. In this study, a Down-Up Sampling Networks (DUS-Net) is proposed to reconstruct high-precision point clouds for 3D anomaly detection by preserving the group center geometric structure. The DUS-Net first introduces a Noise Generation module to generate noisy patches, which facilitates the diversity of training data and strengthens the feature representation for reconstruction. Then, a Down-sampling Network~(Down-Net) is developed to learn an anomaly-free center point cloud from patches with noise injection. Subsequently, an Up-sampling Network~(Up-Net) is designed to reconstruct high-precision point clouds by fusing multi-scale up-sampling features. Our method leverages group centers for construction, enabling the preservation of geometric structure and providing a more precise point cloud. Extensive experiments demonstrate the effectiveness of our proposed method, 
achieving state-of-the-art~(SOTA) performance, with an Object-level AUROC of 79.9\% and 79.5\% and a Point-level AUROC of 71.2\% and 84.7\% on the Real3D-AD and Anomaly-ShapeNet datasets, respectively.
\end{abstract}

\maketitle

\section{Introduction}
\begin{figure}
    \centering
    \includegraphics[width=0.97\linewidth]{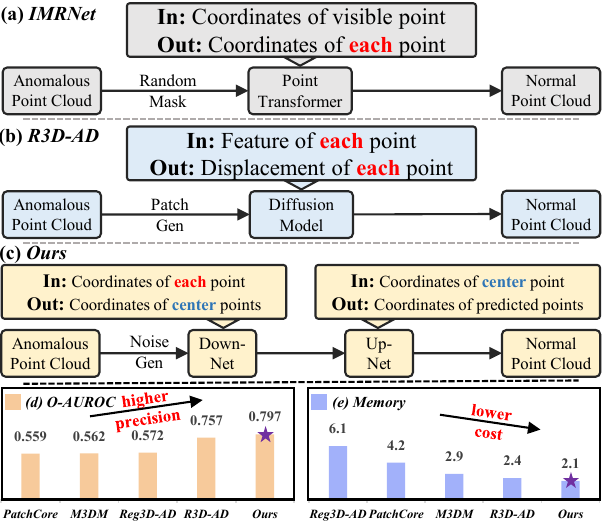}
    \caption{Comparison between previous methods and ours. 
    The previous methods IMRNet~\cite{IMRNet} (a) and R3DAD~\cite{R3DAD} (b) predict the exact coordinates of \textcolor[rgb]{1, 0, 0}{each} point, and our proposed (c) reconstruct a high-precision point cloud from precisely predicted group \textcolor[rgb]{0, .439, .753}{center} points. This leads to higher performance (d) and faster inference speed (e).}
    \label{solution}
\end{figure}
3D anomaly detection (AD) has attracted the spotlight of recent research due to the potential for high-precision industrial product inspection. The task requires the identification of points or regions that deviate from the normal distribution in a given 3D point cloud data. In real-world industrial applications, due to collecting normal samples is easy and labeling anomalous samples is time-consuming and laborious, unsupervised methods that learn only from normal samples are widely used. Recent methods are roughly classified into feature embedding and feature reconstruction.

Feature embedding-based methods use pre-trained models or mathematical descriptors to extract features from normal samples and create a memory bank to match the features of the test samples, where points with large match errors are considered as anomalies. Some methods like BTF~\cite{BTF}, AST~\cite{AST}, M3DM~\cite{M3DM}, and 3D-ADNAS~\cite{3D-ADNAS} achieved excellent performance on MvTec3D-AD datasets~\cite{Mvtec3DAD}. Moreover, by exploiting robust feature extractors, some methods like Reg3D-AD~\cite{Real3D-AD}, Group3AD~\cite{Group3AD}, and ISMP~\cite{ISMP} also attained impressive performance on more challenging datasets such as Anomaly-ShapeNet~\cite{IMRNet} and Real3D-AD~\cite{Real3D-AD}. Nevertheless, these feature embedding methods may have limitations in accurately extracting and representing features for each point.

The feature reconstruction methods encode the point cloud data into a latent space and decode into reconstruction feature space or coordinate space, with high reconstruction errors indicating the presence of anomalies. Some methods based on features, such as Shape-Guided~\cite{Shape-Guided} and CFM~\cite{cfm}, exhibited their effectiveness in distinguishing between normality and abnormality. On the other hand, some point-based accurate reconstruction methods, like IMRNet~\cite{IMRNet} and R3D-AD~\cite{R3DAD}, further improved the ability to detect point-level anomalies by predicting the coordinates or displacements of points. 
However, when handling high-precision point clouds, directly predicting point-level coordinates or displacements may face great challenges because of the large scale of points and complex structure. Grouping high-precision point clouds is a common technique used in recent methods, and the grouping centers maintain the main geometric structure of the point cloud while having a very small scale. Intuitively, group center preserving reconstruction is one of the potential alternatives, which maintains the overall geometric structure and also allows for up-sampling to restore high-precision point clouds. The differences between our solution and previous methods are shown in Figure \ref{solution}.


Motivated by the above facts, we propose a Down-Up Sampling Networks (DUS-Net). Specifically, we first introduce a Noise Generation to generate noisy patches. Instead of directly predicting point-level information, we present a Down-sampling Network~(Down-Net) to predict anomaly-free group centers from noisy patches to maintain overall geometric structures. Then, to meet the demand for high-precision anomaly detection, we design an Upsampling Network~(Up-Net) to reconstruct high-precision normal point clouds by fusing multi-scale up-sampling features. The main contributions of this study are summarized as follows:

\begin{enumerate}
    \item To perform anomaly detection on high-precision point clouds, we propose a Down-Up Sampling Networks (DUS-Net), which allows for reconstructing high-precision and anomaly-free point clouds by capitalizing on group center-level points.

    \item To keep the overall geometric structure of the point clouds, we introduce a Down-Net to predict the group centers from noisy patches generated by a noise generation module, which provides the ability of denoising and also preserves the accurate topology structure points for reconstruction.      

    \item To reconstruct high-precision point clouds, we present an UpSampling Net~(Up-Net) to form multi-scale representations from down-sampled group center point clouds for reconstruction, which fuses multi-scale feature information and enable the generation of high-precision anomaly-free point clouds.
    
    \item Comparative experiments show that our method outperforms previous approaches and achieves SOTA performance, with an Object-level AUROC of 79.9\% and 79.5\% and a Point-level AUROC of 71.2\% and 84.7\% on Real3D-AD and Anomaly-ShapeNet, respectively.
\end{enumerate}

\section{Related Work}
\label{relatedwork}
\subsection{2D Anomaly Detection}
2D anomaly detection is one of the crucial vision tasks, with the objective of detecting and localizing anomalies from images or patches~\cite{Lu2023hvq, Zhao2023omnial}. Current 2D anomaly detection methods predominantly follow two distinct paradigms: generative and discriminative methods~\cite{PangSC2021}. The former leverage neural models such as Autoencoder~\cite{Liu_2025, Tao_2022}, Generative Adversarial Networks~\cite{Yan_Zhang_Xu_Hu_Heng_2021}, and Diffusion~\cite{Wyatt_2022_CVPR} to capture representations or distributions of normal samples and identify anomalies by comparing with a memory bank~\cite{Patchcore,pni} or by reconstructing images with errors~\cite{ZAVRTANIK2021reconstruction}. Methods like SPADE~\cite{spade}, DRAEM~\cite{Zavrtanik2021DRAEM}, and PatchCore~\cite{Patchcore} have achieved very impressive detection results on publicly available datasets. The latter employs supervised learning mechanisms and focuses on training classifiers to distinguish between defective and normal images with the aid of labeled images. Recent approaches such as DRA~\cite{Ding2022catch}, BGAD~\cite{Zhang2023prototypical}, and AHL~\cite{Zhu2024openset} validate the advantages of supervised information in enhancing detection precision. Notably, directly migrating 2D anomaly detection methods to 3D point clouds may encounter significant challenges due to inherent dimensional and structural disparities between data modalities.

\subsection{3D Anomaly Detection}
The 3D Anomaly Detection aims to detect and locate anomalous points or areas within 3D point cloud data~\cite{IMRNet, ISMP, Rani_2024}. Existing methods for 3D anomaly detection can be classified into two main categories: feature embedding and reconstruction methods.

Feature embedding methods employ pre-trained models or mathematical descriptors to extract representative features of normal samples for forming a memory bank, and anomalies are detected by comparing the features of the test sample with those in the memory bank.
Reg3D-AD~\cite{Real3D-AD} used PointMAE~\cite{pointmae} to extract registered features by RANSAC~\cite{ransac} and created a dual memory bank with patch features and point coordinates to compute point anomaly scores during inference. Group3AD~\cite{Group3AD} introduced a contrastive learning approach that maps different groups into distinct clusters to extracts group-level features for better anomaly detection. Looking3D~\cite{Bhunia2024look}, CPMF~\cite{CPMF}, and ISMP~\cite{ISMP} extracted more discriminating features from generated 2D modalities as additional information to improve detection.
Moreover, some multi-modal methods, such as BTF~\cite{BTF}, M3DM~\cite{M3DM}, Shape-Guided~\cite{Shape-Guided}, CFM~\cite{cfm}, and 3D-ADNAS~\cite{3D-ADNAS}, achieved better anomaly detection by fusing features from point clouds and RGB images. Recently, the anomaly detection approaches based on the Large Language Model~(LLM) also showed good detection performance on the zero-shot task~\cite{pointad,wang20253dzal,tang2025exploringpotentialencoderfreearchitectures,xu2025zeroshotanomalydetectionreasoning,wang2024zeroshot3danomalylocalization}. 

Reconstruction methods encode the point cloud into a latent space and decode it back to its original form, where points with high reconstruction errors are regarded as anomalies. IMRNet~\cite{IMRNet} randomly masked and predicted the corresponding invisible coordinates to detect anomalies. R3D-AD~\cite{R3DAD} employed the diffusion model to compute the displacement of points for anomaly detection. Moreover, Splatpose~\cite{splat} and Splatpose++~\cite{splat++} used the Gaussian splatting to achieve multi-view reconstruction and anomaly detection. Although these methods achieve promising results, they may face great challenges in high-precision point clouds, which exhibit the characteristics of large scale and high complexity. 
\section{Approach}  
\begin{figure*}[tt]
    \centering
\includegraphics[width=\linewidth]{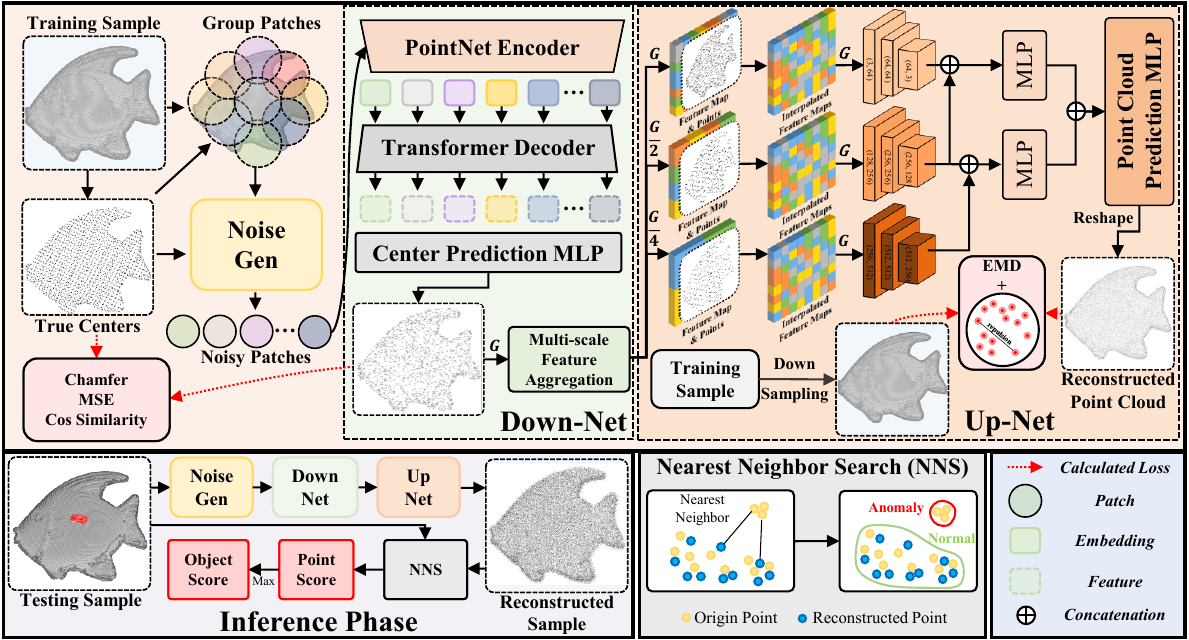}
    \caption{The pipeline of the proposed DUS-Net. The input point cloud is grouped by farthest point sampling (FPS) and K-nearest neighbor search, followed by noise injection using the Noise-Gen module. Then, the Down-Net is trained to predict anomaly-free center points from noisy patches. Finally, the Up-Net is used to reconstruct the predicted center point cloud into a high-precision anomaly-free point cloud by fusing multi-scale features, and points with large reconstruction errors are considered as anomalies.}
    \label{pipeline}
\end{figure*}

Reconstruction-based methods ~\cite{IMRNet, R3DAD} have shown their effectiveness in 3D anomaly detection. Nevertheless, when confronted with high-precision point clouds, these methods may have limitations in reconstruction quality and computational efficiency. To address these challenges, we propose the Down-Up Sampling Networks (DUS-Net) to reconstruct high-precision point clouds by preserving the group centers, and the overall framework is illustrated in Figure \ref{pipeline}. It consists of three components: a Noise generation module, a center-preserved down-sampling module, and a multi-scale up-sampling module. Each component is described in detail in the following sections.

\subsection{Patch Augmentation with Noise Generation}
\label{Noise-Gen}
For 3D anomaly detection, the available training data is often very limited. To enrich the training data, some methods rely on generating pseudo-anomalies to improve their representation and detection capability. Nevertheless, the quality of the generated anomalies is difficult to guarantee, thereby affecting the performance. To address this problem, we propose a noise generation module (Noise-Gen) to diversify the data and promote the model to learn the ability of reconstructing normal points from noisy ones.

 
Given a point cloud $\mathcal{P}$ $\in$ $\mathbb{R}^{N\times3}$, we partition it into $G$ patches $\{P_i\}_{i=1}^G$ via farthest point sampling~(FPS) and K-nearest neighbors~(KNN). Each patch $P_i$ $\in$ $\mathbb{R}^{(K+1)\times3}$ is composed by $P_i$ = $\{c_i \cup \mathcal{N}_K(c_i)\}$, where $c_i$ denotes the center point selected by the FPS, and $\mathcal{N}_K(c_i)$ means the $K$ nearest neighbors of $c_i$. To diversify the point cloud, Gaussian noises are injected into each group with two controllable noise parameters $\alpha$ and $\beta$. The patch after noise injection can be formulated as
\begin{equation}
\begin{aligned}
    \tilde{P}_i &= GSN(P_i, \alpha, \beta),\\
    &= \{GSN(c_i, \alpha)  \cup  GSN(\mathcal{N}_K(c_i), \beta)\},\\
    &= \{\tilde{c}_i \cup \tilde{\mathcal{N}}_K(c_i)\},
\end{aligned}
\end{equation}
where $GSN(\cdot, \cdot)$ denotes the operator of injecting at a certain level of Gaussian noise on a point or a set of points, and $\tilde{c}_i$ and $\tilde{\mathcal{N}}_K(c_i)$ represent the center and its nearest neighbors after injection of Gaussian noise.

After the noise injection process, points within each group are subjected to different levels of Gaussian noise. Despite causing subtle distortions to the structure, this process enriches the point cloud and facilitates a robust representation for downstream tasks.

\subsection{Center-preserved Down-sampling with Down-Net}
\label{DownNet}
Point cloud data exhibits the characteristics of large scale and complexity. Directly reconstructing or predicting each point may face great challenges. To precisely reconstruct the point cloud, we propose a Down-Net to extract robust representations from noisy patches while preserving group centers.

Specifically, the noisy patches after injecting Gaussian noise are input into a Point-Net~\cite{pointnet} to extract key features. While their centers are encoded as position embeddings by using a multilayer perceptron~(MLP) to avoid leakage of the center information. These processes can be formalized as
\begin{equation}
\begin{aligned}
    E_c&=MLP(\tilde{c}_i),\ E_c\in\mathbb{R}^{G\times C_1};\\
    E_p&=PointNet(\mathcal{\tilde{N}}_K(c_i)), \ E_p\in\mathbb{R}^{G\times C_2}.
\end{aligned}
\end{equation}
where $MLP$ and $PointNet$ mean the feature extractor MLP and Point-Net, respectively, $C_1$ and $C_2$ denote the dimensions of the position embeddings and patch feature embeddings, respectively.

These patch embeddings, along with the position embeddings, are then fed into a transformer-architecture decoder to learn group-level features, where the attention mechanism captures the relationship of different patches to assist the decoding of each patch. The decoding process can be formalized as
\begin{equation}
    E_f=Decoder(Concat(E_c, E_p)),\ E_f\in\mathbb{R}^{G\times C_3},
\end{equation}
where $Concat$ denotes the concatenation operator of different feature embeddings, $Decoder$ represents the decoder process, and $C_3$ stands for the dimension of the decoded patch features.

Finally, an MLP is used to predict centers, followed by a reshape operation to restore the predicted centers into the form of a 3D point cloud. The center prediction process can be represented as
\begin{equation}
    \tilde{\mathcal{P}}_c = Reshape(MLP(E_f)), \tilde{\mathcal{P}}_c \in \mathbb{R}^{G\times3}.
\end{equation}

To precisely predict all centers under noise conditions, the Mean Squared Error~(MSE) loss is used to minimize the errors between the original and predicted centers and to promote the learning of a robust feature representation for the group center geometric structure of the point cloud~\cite{ren2022balancedmseimbalancedvisual}, which can be defined as
\begin{equation}
\mathcal{L}_{MSE}(\mathcal{P}_c, \tilde{\mathcal{P}}_c) = \frac{1}{N} \sum_{p_i \in \mathcal{P}_c} \sum_{p_j \in \tilde{\mathcal{P}}_c} \parallel p_i - p_j \parallel^2_2,
\end{equation}
where $\mathcal{P}_c$ and $\tilde{\mathcal{P}}_c$ denote the original and predicted group center set, respectively, and $\parallel \cdot \parallel$ represents the 2-norm between two points.

To keep directional consistency on the predicted centers, the cosine similarity loss is further introduced, which is defined as
\begin{equation}
\mathcal{L}_{COS}(\mathcal{P}_c, \tilde{\mathcal{P}}_c) = \frac{1}{N} \sum_{p_i \in \mathcal{P}_c} \sum_{p_j \in \tilde{\mathcal{P}}_c} \left( 1 - \frac{p_i \cdot p_j}{\parallel p_i \parallel_2 \parallel p_j \parallel_2} \right),
\end{equation}

Moreover, the Chamfer distance loss~\cite{wu2021densityawarechamferdistancecomprehensive} is employed to ensure macro-scale structural coherence, which is defined as
\begin{equation}
\begin{aligned}
\mathcal{L}_{CD}(\mathcal{P}_c, \tilde{\mathcal{P}}_c) &= \frac{1}{\mathcal{P}_c} \sum_{p_i \in \mathcal{P}_c} \min_{p_j \in \tilde{\mathcal{P}}_c} \left\| p_i - p_j \right\|_2^2 \\
&+ \frac{1}{\tilde{\mathcal{P}}_c} \sum_{p_j \in \tilde{\mathcal{P}}_c} \min_{p_i \in \mathcal{P}_c} \left\| p_j - p_i \right\|_2^2.
\end{aligned}
\end{equation}

The Chamfer distance adopts a bidirectional matching manner, with the first term ensuring the existence of corresponding predicted centers for each original center and the second term keeping the proximity of each predicted center to the original centers.
Thus, the overall loss to optimize the Down-Net can be expressed as

\begin{equation}
\label{down_loss}
\mathcal{L}_{Down} = \mathcal{L}_{MSE} + \mathcal{L}_{COS} + \mathcal{L}_{CD}.
\end{equation}

\subsection{Multiscale Reconstruction with Up-Net}
\label{UpNet}
Down-Net aims to predict centers for noisy patches to keep the main geometric structure of a point cloud. While these center-level points are not sufficient for high-precision anomaly detection. To address this problem, we propose an Up-Net to reconstruct high-precision point clouds by fusing multi-scale feature representations.

Specifically, the anomaly-free center-level point cloud $\tilde{\mathcal{P}}_c$ predicted by Down-Net is further processed by the feature aggregation technique in PointNet++~\cite{pointnet2} to extract multi-scale features from the down-sampled point clouds with the scale of $1$, $1/2$, and $1/4$, respectively. Formally, the features extracted in the $l$-th scale can be defined as~\cite{pointnet2}
\begin{equation}
\begin{aligned}
 F_i^{l} &= \mathop{MaxPool}\limits_{p_j^{l-1} \in \mathcal{N}^{l-1}(p_i^{l})} \left(MLP \big([p_j^{l-1} - p_i^{l};\, F_j^{l-1}]\big) \right), 1 \le l \le 2,
\end{aligned}
\end{equation}
where $p_i^{l}$ denotes a point in the $l$-th scale down-sampled point cloud $\tilde{\mathcal{P}}^l$, $p_j^{l-1}$ and $F_j^{l-1}$ represents the points from the neighbor set $\mathcal{N}^{l-1}(p_i^{l})$ of $p_i^{l}$ in the $l-1$-th scale and the corresponding extracted feature, respectively, and the symbols $MaxPool$ and $MLP$ stand for the operator of max-pooling and the feature extractor MLP, respectively.

Subsequently, the learned feature maps in the lower scales are interpolated up to align the features in the original scale by weighting the features of three nearest neighbors, which can be formalized as
\begin{equation}
\begin{aligned}
     \tilde{F}^{l}&=TriInter(\tilde{\mathcal{P}}_c, \tilde{\mathcal{P}}^l, F^{l}),\\
     \tilde{F}^{l}_i &= \frac{\sum_{j=1}^3 w_i^j \cdot F^{l}_j}{\sum_{j=1}^3 w_i^j}, p_i \in \tilde{\mathcal{P}}_c /\tilde{\mathcal{P}}^l,\\
    w_i^j &= \frac{1}{\|p_i - p_j\|_2},
\end{aligned}
\end{equation}
where $TriInter$ means the triple nearest neighbor interpolation, $p_i$ denotes the point that appear in the original scale point cloud $\tilde{\mathcal{P}}_c$ but not in the current $l$ scale point cloud $\tilde{\mathcal{P}}^l$, $p_j$ represent the $j$-th nearest neighbor of $p_i$ in $\tilde{\mathcal{P}}^l$, and $F_j^{l-1}$ stands for the feature of the point $p_j$ in the $l$-th scale.

These aligned features are separately processed by a 3-layer convolution with a kernel size of $1\times1$. To promote the fusion of features, features in the near scales are concatenated and fed into an MLP to reduce the feature dimensions by half. These two-branch intermediate features are further concatenated for MLPs to predict points with the dimension size of $(G, 3 \times \gamma)$, and these reconstructed points are finally reshaped to an up-sampled high-precision point cloud with the dimension size of $(G \times \gamma, 3)$.
This process can be formalized as
\begin{equation}
\begin{aligned}
      \hat{\mathcal{P}} = &Reshape\bigg(MLPs\Big(Concat\big(MLP\big(Concat(Convs\small(\tilde{\mathcal{P}}_c \small), \\ &Convs\small(\tilde{F}^1\small))\big), MLP\big(Concat(Convs\small(\tilde{F}^1\small), Convs\small(\tilde{F}^2\small))\big) \big)\Big)\bigg). 
\end{aligned}
\end{equation}

To train the Up-Net, the repulsion loss~\cite{yu2018punetpointcloudupsampling} is used to ensure the generated point cloud more evenly, which is defined as
\begin{equation}
    \mathcal{L}_{REP}(\hat{\mathcal{P}}) = \sum_{p_i\in\hat{\mathcal{P}}} \sum_{j=1}^K \eta(\|p_{i} - p_{j}\|_2) \omega(\|p_i - p_j\|_2),
\end{equation}
where $K$ is the number of neighbors, $\eta(d) = -d$ denotes the repulsion term to penalize the affinity between neighboring points, and  $\omega(d) = e^{-d^2}$ represents the fast-decaying weight function for the repulsion term.

Moreover, Earth Mover’s distance~(EMD) loss is introduced to preserve the alignment of point clouds in shape and density distribution by forcing point-to-point matching, which is defined as
\begin{equation}
    \mathcal{L}_{EMD}(\hat{\mathcal{P}}, \mathcal{P}_{G}) = \sum_{p_i \in \hat{\mathcal{P}}} \min_{p_j\in\mathcal{P}_{G}}\left\| p_i - p_j \right\|_2,
\end{equation}
where $\mathcal{P}_{G}$ denotes the ground truth high-precision point cloud. Thus, the overall loss to optimize the Up-Net can be formalized as
\begin{equation}\mathcal{L}_{Up}=\mathcal{L}_{REP}+\mathcal{L}_{EMD}. \label{up_loss}
\end{equation}

\subsection{Training and Inference}
\textbf{Training.} The proposed Down-Net and Up-Net are trained separately. Firstly, the Down-Net is trained on noisy patches generated by the Noise-Gen module using the loss $\mathcal{L}_{Down}$ in \ref{down_loss}. The Down-Net is then frozen and employed to convert raw point clouds into center prediction results. These intermediate representations are subsequently used to optimize Up-Net with the loss $\mathcal{L}_{Up}$ in \ref{up_loss}.  

\textbf{Inference.} During the testing phase, the testing point cloud $\mathcal{P}_{in}$ after noise injection by Noise-Gen is reconstructed into an anomaly-free point cloud $\mathcal{P}_{out}$ using the trained Down-Net and Up-Net. Although the precision of the output point cloud $P_{out}$ is high, the number of its points may not exactly match the raw point cloud. Thus, the nearest neighbor distance between two point clouds is calculated as the point anomaly score:
\begin{equation}
s_i = \min_{p_j \in \mathcal{P}_{out}} \left\| p_i - p_j \right\|_2, p_{i} \in \mathcal{P}_{in}.
\end{equation}

To facilitate pixel-level anomaly detection, the anomaly score is normalized using a weighting method~\cite{Patchcore}, which is denoted as:
\begin{equation}
\tilde{s}_{i} = \left( 1 - \frac{\exp \left\| p_i - p_j \right\|_2}{\sum_{p_{j} \in \mathcal{N}_3(p_i, \mathcal{P}_{out})} \exp \left\| p_i - p_j \right\|_2} \right) s_i,
\end{equation}
where $p_{j} \in \mathcal{N}_3(p_i, \mathcal{P}_{out})$ denotes the 3 nearest points of $p_i$ in $P_{out}$. Further, with the maximum value of the point-level anomaly scores, the object-level score $S$ is calculated as
\begin{equation}
    S= \max_{p_i \in \mathcal{P}_{in}}(\tilde{s}_{i})
\end{equation}

 \begin{table*}[!ht]
  \centering
  \caption{O-AUROC performance of different methods on Real3D-AD across 40 categories, where best and second-place results are highlighted in \textcolor[rgb]{1, 0, 0}{\textbf{red}} and \textcolor[rgb]{0, .439, .753}{\textbf{blue}}, respectively.}
  \resizebox{0.95\textwidth}{!}{
    \begin{tabular}{c|cccccccccccccc}
    \toprule
    \multicolumn{15}{c}{\textbf{O-AUROC}} \\
    \midrule
    \textbf{Method} & \textbf{ashtray0} & \textbf{bag0} & \textbf{bottle0} & \textbf{bottle1} & \textbf{bottle3} & \textbf{bowl0} & \textbf{bowl1} & \textbf{bowl2} & \textbf{bowl3} & \textbf{bowl4} & \textbf{bowl5} & \textbf{bucket0} & \textbf{bucket1} & \textbf{cap0} \\
    \midrule
    \textbf{BTF(Raw) (CVPR23’)} & 0.578  & 0.410  & 0.597  & 0.510  & 0.568  & 0.564  & 0.264  & 0.525  & 0.385  & 0.664  & 0.417  & 0.617  & 0.321  & 0.668  \\
    \textbf{BTF(FPFH) (CVPR23’)} & 0.420  & 0.546  & 0.344  & 0.546  & 0.322  & 0.509  & 0.668  & 0.510  & 0.490  & 0.609  & 0.699  & 0.401  & 0.633  & 0.618  \\
    \textbf{M3DM (CVPR23’)} & 0.577  & 0.537  & 0.574  & 0.637  & 0.541  & 0.634  & 0.663  & 0.684  & 0.617  & 0.464  & 0.409  & 0.309  & 0.501  & 0.557  \\
    \textbf{PatchCore(FPFH) (CVPR22’)} & 0.587  & 0.571  & 0.604  & 0.667  & 0.572  & 0.504  & 0.639  & 0.615  & 0.537  & 0.494  & 0.558  & 0.469  & 0.551  & 0.580  \\
    \textbf{PatchCore(PointMAE) (CVPR22')} & 0.591  & 0.601  & 0.513  & 0.601  & 0.650  & 0.523  & 0.629  & 0.458  & 0.579  & 0.501  & 0.593  & 0.593  & 0.561  & 0.589  \\
    \textbf{CPMF (PR24’)} & 0.353  & 0.643  & 0.520  & 0.482  & 0.405  & 0.783  & 0.639  & 0.625  & 0.658  & 0.683  & 0.685  & 0.482  & 0.601  & 0.601  \\
    \textbf{Reg3D-AD (NeurIPS23’)} & 0.597  & \textcolor[rgb]{ 0,  .439,  .753}{\textbf{0.706 }} & 0.486  & 0.695  & 0.525  & 0.671  & 0.525  & 0.490  & 0.348  & 0.663  & 0.593  & 0.610  & 0.752  & 0.693  \\
    \textbf{IMRNet (CVPR24’)} & 0.671  & 0.660  & 0.552  & 0.700  & 0.640  & 0.681  & 0.702  & 0.685  & 0.599  & 0.676  & \textcolor[rgb]{ 0,  .439,  .753}{\textbf{0.710 }} & 0.580  & \textcolor[rgb]{ 0,  .439,  .753}{\textbf{0.771 }} & 0.737  \\
    \textbf{R3D-AD (ECCV24’)} & \textcolor[rgb]{ 0,  .439,  .753}{\textbf{0.833 }} & \textcolor[rgb]{ 1,  0,  0}{\textbf{0.720 }} & \textcolor[rgb]{ 0,  .439,  .753}{\textbf{0.733 }} & \textcolor[rgb]{ 0,  .439,  .753}{\textbf{0.737 }} & \textcolor[rgb]{ 0,  .439,  .753}{\textbf{0.781 }} & \textcolor[rgb]{ 0,  .439,  .753}{\textbf{0.819 }} & \textcolor[rgb]{ 0,  .439,  .753}{\textbf{0.778 }} & \textcolor[rgb]{ 0,  .439,  .753}{\textbf{0.741 }} & \textcolor[rgb]{ 0,  .439,  .753}{\textbf{0.767 }} & \textcolor[rgb]{ 0,  .439,  .753}{\textbf{0.744 }} & 0.656  & \textcolor[rgb]{ 0,  .439,  .753}{\textbf{0.683 }} & 0.756  & \textcolor[rgb]{ 0,  .439,  .753}{\textbf{0.822 }} \\
    \textbf{DU-Net (Ours)} & \textcolor[rgb]{ 1,  0,  0}{\textbf{0.867 }} & 0.605  & \textcolor[rgb]{ 1,  0,  0}{\textbf{0.838 }} & \textcolor[rgb]{ 1,  0,  0}{\textbf{0.871 }} & \textcolor[rgb]{ 1,  0,  0}{\textbf{0.827 }} & \textcolor[rgb]{ 1,  0,  0}{\textbf{0.844 }} & \textcolor[rgb]{ 1,  0,  0}{\textbf{0.869 }} & \textcolor[rgb]{ 1,  0,  0}{\textbf{0.952 }} & \textcolor[rgb]{ 1,  0,  0}{\textbf{0.839 }} & \textcolor[rgb]{ 1,  0,  0}{\textbf{0.859 }} & \textcolor[rgb]{ 1,  0,  0}{\textbf{0.776 }} & \textcolor[rgb]{ 1,  0,  0}{\textbf{0.838 }} & \textcolor[rgb]{ 1,  0,  0}{\textbf{0.822 }} & \textcolor[rgb]{ 1,  0,  0}{\textbf{0.859 }} \\
    \midrule

    \midrule
    \textbf{Method} & \textbf{cap3} & \textbf{cap4} & \textbf{cap5} & \textbf{cup0} & \textbf{cup1} & \textbf{eraser0} & \textbf{headset0} & \textbf{headset1} & \textbf{helmet0} & \textbf{helmet1} & \textbf{helmet2} & \textbf{helmet3} & \textbf{jar0} & \textbf{micro.} \\
    \midrule
    \textbf{BTF(Raw)(CVPR23’)} & 0.527  & 0.468  & 0.373  & 0.403  & 0.521  & 0.525  & 0.378  & 0.515  & 0.553  & 0.349  & 0.602  & 0.526  & 0.420  & 0.563  \\
    \textbf{BTF(FPFH) (CVPR23’)} & 0.522  & 0.520  & 0.586  & 0.586  & 0.610  & \textcolor[rgb]{ 0,  .439,  .753}{\textbf{0.719 }} & 0.520  & 0.490  & 0.571  & 0.719  & 0.542  & 0.444  & 0.424  & 0.671  \\
    \textbf{M3DM(CVPR23’)} & 0.423  & \textcolor[rgb]{ 1,  0,  0}{\textbf{0.777 }} & 0.639  & 0.539  & 0.556  & 0.627  & 0.577  & 0.617  & 0.526  & 0.427  & 0.623  & 0.374  & 0.441  & 0.357  \\
    \textbf{PatchCore(FPFH)(CVPR22’)} & 0.453  & \textcolor[rgb]{ 0,  .439,  .753}{\textbf{0.757 }} & \textcolor[rgb]{ 1,  0,  0}{\textbf{0.790 }} & 0.600  & 0.586  & 0.657  & 0.583  & 0.637  & 0.546  & 0.484  & 0.425  & 0.404  & 0.472  & 0.388  \\
    \textbf{PatchCore(PointMAE) (CVPR22')} & 0.476  & 0.727  & 0.538  & 0.610  & 0.556  & 0.677  & 0.591  & 0.627  & 0.556  & 0.552  & 0.447  & 0.424  & 0.483  & 0.488  \\
    \textbf{CPMF(PR24’)} & 0.551  & 0.553  & 0.697  & 0.497  & 0.499  & 0.689  & 0.643  & 0.458  & 0.555  & 0.589  & 0.462  & 0.520  & 0.610  & 0.509  \\
    \textbf{Reg3D-AD(NeurIPS23’)} & 0.725  & 0.643  & 0.467  & 0.510  & 0.538  & 0.343  & 0.537  & 0.610  & 0.600  & 0.381  & 0.614  & 0.367  & 0.592  & 0.414  \\
    \textbf{IMRNet(CVPR24’)} & \textcolor[rgb]{ 1,  0,  0}{\textbf{0.775 }} & 0.652  & 0.652  & 0.643  & \textcolor[rgb]{ 0,  .439,  .753}{\textbf{0.757 }} & 0.548  & 0.720  & 0.676  & 0.597  & 0.600  & \textcolor[rgb]{ 0,  .439,  .753}{\textbf{0.641 }} & 0.573  & \textcolor[rgb]{ 0,  .439,  .753}{\textbf{0.780 }} & 0.755  \\
    \textbf{R3D-AD(ECCV24’)} & \textcolor[rgb]{ 0,  .439,  .753}{\textbf{0.730 }} & 0.681  & 0.670  & \textcolor[rgb]{ 0,  .439,  .753}{\textbf{0.776 }} & \textcolor[rgb]{ 0,  .439,  .753}{\textbf{0.757 }} & \textcolor[rgb]{ 1,  0,  0}{\textbf{0.890 }} & \textcolor[rgb]{ 0,  .439,  .753}{\textbf{0.738 }} & \textcolor[rgb]{ 0,  .439,  .753}{\textbf{0.795 }} & \textcolor[rgb]{ 1,  0,  0}{\textbf{0.757 }} & \textcolor[rgb]{ 0,  .439,  .753}{\textbf{0.720 }} & 0.633  & \textcolor[rgb]{ 0,  .439,  .753}{\textbf{0.707 }} & \textcolor[rgb]{ 1,  0,  0}{\textbf{0.838 }} & \textcolor[rgb]{ 0,  .439,  .753}{\textbf{0.762 }} \\
    \textbf{DU-Net (Ours)} & \textcolor[rgb]{ 1,  0,  0}{\textbf{0.775 }} & 0.736  & \textcolor[rgb]{ 0,  .439,  .753}{\textbf{0.739 }} & \textcolor[rgb]{ 1,  0,  0}{\textbf{0.869 }} & \textcolor[rgb]{ 1,  0,  0}{\textbf{0.776 }} & 0.644  & \textcolor[rgb]{ 1,  0,  0}{\textbf{0.741 }} & \textcolor[rgb]{ 1,  0,  0}{\textbf{0.961 }} & \textcolor[rgb]{ 0,  .439,  .753}{\textbf{0.743 }} & \textcolor[rgb]{ 1,  0,  0}{\textbf{0.860 }} & \textcolor[rgb]{ 1,  0,  0}{\textbf{0.841 }} & \textcolor[rgb]{ 1,  0,  0}{\textbf{0.859 }} & 0.744  & \textcolor[rgb]{ 1,  0,  0}{\textbf{0.837 }} \\
    \midrule
    \midrule
    \textbf{Method} & \textbf{shelf0} & \textbf{tap0} & \textbf{tap1} & \textbf{vase0} & \textbf{vase1} & \textbf{vase2} & \textbf{vase3} & \textbf{vase4} & \textbf{vase5} & \textbf{vase7} & \textbf{vase8} & \textbf{vase9} & \textbf{Average} & \textbf{Ranking} \\
    \midrule
    \textbf{BTF(Raw)(CVPR23’)} & 0.164  & 0.525  & 0.573  & 0.531  & 0.549  & 0.410  & 0.717  & 0.425  & 0.585  & 0.448  & 0.424  & 0.564  & 0.493  & 7.700  \\
    \textbf{BTF(FPFH) (CVPR23’)} & 0.609  & 0.560  & 0.546  & 0.342  & 0.219  & 0.546  & 0.699  & 0.510  & 0.409  & 0.518  & 0.668  & 0.268  & 0.528  & 7.025  \\
    \textbf{M3DM(CVPR23’)} & 0.564  & \textcolor[rgb]{ 1,  0,  0}{\textbf{0.754 }} & 0.739  & 0.423  & 0.427  & 0.737  & 0.439  & 0.476  & 0.317  & 0.657  & 0.663  & \textcolor[rgb]{ 0,  .439,  .753}{\textbf{0.663 }} & 0.552  & 6.800  \\
    \textbf{PatchCore(FPFH) (CVPR22’)} & 0.494  & \textcolor[rgb]{ 0,  .439,  .753}{\textbf{0.753 }} & 0.766  & 0.455  & 0.423  & 0.721  & 0.449  & 0.506  & 0.417  & 0.693  & 0.662  & 0.660  & 0.568  & 6.300  \\
    \textbf{PatchCore(PointMAE) (CVPR22')} & 0.523  & 0.458  & 0.538  & 0.447  & 0.552  & \textcolor[rgb]{ 0,  .439,  .753}{\textbf{0.741 }} & 0.460  & 0.516  & 0.579  & 0.650  & 0.663  & 0.629  & 0.562  & 6.325  \\
    \textbf{CPMF(PR24’)} & 0.685  & 0.359  & 0.697  & 0.451  & 0.345  & 0.582  & 0.582  & 0.514  & 0.618  & 0.397  & 0.529  & 0.609  & 0.559  & 6.350  \\
    \textbf{Reg3D-AD(NeurIPS23’)} & \textcolor[rgb]{ 0,  .439,  .753}{\textbf{0.688 }} & 0.676  & 0.641  & 0.533  & 0.702  & 0.605  & 0.650  & 0.500  & 0.520  & 0.462  & 0.620  & 0.594  & 0.572  & 6.400  \\
    \textbf{IMRNet(CVPR24’)} & 0.603  & 0.676  & 0.696  & 0.533  & \textcolor[rgb]{ 0,  .439,  .753}{\textbf{0.757 }} & 0.614  & 0.700  & 0.524  & 0.676  & 0.635  & 0.630  & 0.594  & 0.661  & 3.925  \\
    \textbf{R3D-AD(ECCV24’)} & \textcolor[rgb]{ 1,  0,  0}{\textbf{0.696 }} & 0.736  & \textcolor[rgb]{ 1,  0,  0}{\textbf{0.900 }} & \textcolor[rgb]{ 0,  .439,  .753}{\textbf{0.788 }} & 0.729  & \textcolor[rgb]{ 1,  0,  0}{\textbf{0.752 }} & \textcolor[rgb]{ 0,  .439,  .753}{\textbf{0.742 }} & \textcolor[rgb]{ 0,  .439,  .753}{\textbf{0.630 }} & \textcolor[rgb]{ 0,  .439,  .753}{\textbf{0.757 }} & \textcolor[rgb]{ 0,  .439,  .753}{\textbf{0.771 }} & \textcolor[rgb]{ 0,  .439,  .753}{\textbf{0.721 }} & \textcolor[rgb]{ 1,  0,  0}{\textbf{0.718 }} & \textcolor[rgb]{ 0,  .439,  .753}{\textbf{0.749 }} & \textcolor[rgb]{ 0,  .439,  .753}{\textbf{2.150 }} \\
    \textbf{DU-Net (Ours)} & \textcolor[rgb]{ 0,  .439,  .753}{\textbf{0.688 }} & 0.739  & \textcolor[rgb]{ 0,  .439,  .753}{\textbf{0.840 }} & \textcolor[rgb]{ 1,  0,  0}{\textbf{0.833 }} & \textcolor[rgb]{ 1,  0,  0}{\textbf{0.808 }} & 0.644  & \textcolor[rgb]{ 1,  0,  0}{\textbf{0.766 }} & \textcolor[rgb]{ 1,  0,  0}{\textbf{0.749 }} & \textcolor[rgb]{ 1,  0,  0}{\textbf{0.838 }} & \textcolor[rgb]{ 1,  0,  0}{\textbf{0.844 }} & \textcolor[rgb]{ 1,  0,  0}{\textbf{0.808 }} & 0.525  & \textcolor[rgb]{ 1,  0,  0}{\textbf{0.797 }} & \textcolor[rgb]{ 1,  0,  0}{\textbf{1.775 }} \\
    \bottomrule
    \end{tabular}%
}
  \label{results1}
\end{table*}

\begin{table*}[!ht]
  \centering
  \caption{O-AUROC performance of different methods on Real3D-AD across 12 categories, where best and second-place results are highlighted in \textcolor[rgb]{1, 0, 0}{\textbf{red}} and \textcolor[rgb]{0, .439, .753}{\textbf{blue}}, respectively.}
  \resizebox{0.95\textwidth}{!}{
    \begin{tabular}{c|cccccccccccccc}
    \toprule
    \multicolumn{14}{c}{\textbf{O-AUROC}}                                                                         &  \\
    \midrule
    \textbf{Method} & \textbf{Airplane} & \textbf{Car} & \textbf{Candy} & \textbf{Chicken} & \textbf{Diamond} & \textbf{Duck} & \textbf{Fish} & \textbf{Gemstone} & \textbf{Seahorse} & \textbf{Shell} & \textbf{Starfish} & \textbf{Toffees} & \textbf{Average} & \textbf{Ranking} \\
    \midrule
    \textbf{BTF(Raw) (CVPR23')} & 0.730  & 0.647  & 0.539  & 0.789  & 0.707  & 0.691  & 0.602  & 0.686  & 0.596  & 0.396  & 0.530  & 0.703  & 0.635  & 6.538  \\
    \textbf{BTF(FPFH) (CVPR23')} & 0.520  & 0.560  & 0.630  & 0.432  & 0.545  & 0.784  & 0.549  & 0.648  & \textcolor[rgb]{ 0,  .439,  .753}{\textbf{0.779 }} & 0.754  & 0.575  & 0.462  & 0.603  & 7.000  \\
    \textbf{M3DM (CVPR23')} & 0.434  & 0.541  & 0.552  & 0.683  & 0.602  & 0.433  & 0.540  & 0.644  & 0.495  & 0.694  & 0.551  & 0.450  & 0.552  & 9.000  \\
    \textbf{PatchCore(FPFH) (CVPR22')} & \textcolor[rgb]{ 1,  0,  0}{\textbf{0.882 }} & 0.590  & 0.541  & \textcolor[rgb]{ 0,  .439,  .753}{\textbf{0.837 }} & 0.574  & 0.546  & 0.675  & 0.370  & 0.505  & 0.589  & 0.441  & 0.565  & 0.593  & 7.692  \\
    \textbf{PatchCore(PointMAE) (CVPR22')} & 0.726  & 0.498  & 0.663  & 0.827  & 0.783  & 0.489  & 0.630  & 0.374  & 0.539  & 0.501  & 0.519  & 0.585  & 0.594  & 7.769  \\
    \textbf{CPMF (PR24')} & 0.701  & 0.551  & 0.552  & 0.504  & 0.523  & 0.582  & 0.558  & 0.589  & 0.729  & 0.653  & 0.700  & 0.390  & 0.586  & 8.000  \\
    \textbf{Reg3D-AD (NeurIPS23')} & 0.716  & 0.697  & 0.685  & \textcolor[rgb]{ 1,  0,  0}{\textbf{0.852 }} & 0.900  & 0.584  & \textcolor[rgb]{ 0,  .439,  .753}{\textbf{0.915 }} & 0.417  & 0.762  & 0.583  & 0.506  & 0.827  & 0.704  & 5.231  \\
    \textbf{IMRNet (CVPR24')} & 0.762  & 0.711  & 0.755  & 0.780  & \textcolor[rgb]{ 0,  .439,  .753}{\textbf{0.905 }} & 0.517  & 0.880  & 0.674  & 0.604  & 0.665  & 0.674  & 0.774  & 0.725  & 4.385  \\
    \textbf{R3D-AD (ECCV24')} & 0.772  & 0.696  & 0.713  & 0.714  & 0.685  & \textcolor[rgb]{ 1,  0,  0}{\textbf{0.909 }} & 0.692  & 0.665  & 0.720  & \textcolor[rgb]{ 1,  0,  0}{\textbf{0.840 }} & 0.701  & 0.703  & 0.734  & 4.000  \\
    \textbf{ISMP (AAAI25')} & \textcolor[rgb]{ 0,  .439,  .753}{\textbf{0.858 }} & \textcolor[rgb]{ 0,  .439,  .753}{\textbf{0.731 }} & \textcolor[rgb]{ 0,  .439,  .753}{\textbf{0.852 }} & 0.714  & \textcolor[rgb]{ 1,  0,  0}{\textbf{0.948 }} & 0.712  & \textcolor[rgb]{ 1,  0,  0}{\textbf{0.945 }} & 0.468  & 0.729  & 0.623  & 0.660  & \textcolor[rgb]{ 1,  0,  0}{\textbf{0.842 }} & \textcolor[rgb]{ 0,  .439,  .753}{\textbf{0.757 }} & \textcolor[rgb]{ 0,  .439,  .753}{\textbf{3.462 }} \\
    \textbf{DU-Net (Ours)} & 0.718  & \textcolor[rgb]{ 1,  0,  0}{\textbf{0.738 }} & \textcolor[rgb]{ 1,  0,  0}{\textbf{0.856 }} & 0.696  & 0.824  & \textcolor[rgb]{ 0,  .439,  .753}{\textbf{0.844}} & 0.908  & \textcolor[rgb]{ 1,  0,  0}{\textbf{0.733}} & \textcolor[rgb]{ 1,  0,  0}{\textbf{0.814}} & \textcolor[rgb]{ 0,  .439,  .753}{\textbf{0.822}} & \textcolor[rgb]{ 0,  .439,  .753}{\textbf{0.755}} & \textcolor[rgb]{ 0,  .439,  .753}{\textbf{0.834}} & \textcolor[rgb]{ 1,  0,  0}{\textbf{0.795}} & \textcolor[rgb]{ 1,  0,  0}{\textbf{2.615}} \\
    \bottomrule
    \end{tabular}%
}
  \label{results3}
\end{table*}

 \begin{table*}[!ht]
  \centering
  \caption{P-AUROC performance of different methods on Anomaly-ShapeNet across 40 categories, where best and second-place results are highlighted in \textcolor[rgb]{1, 0, 0}{\textbf{red}} and \textcolor[rgb]{0, .439, .753}{\textbf{blue}}, respectively.}
  \resizebox{0.95\textwidth}{!}{
    \begin{tabular}{c|cccccccccccccc}
    \toprule
    \multicolumn{15}{c}{\textbf{P-AUROC}} \\
    \midrule
    \textbf{Method} & \textbf{ashtray0} & \textbf{bag0} & \textbf{bottle0} & \textbf{bottle1} & \textbf{bottle3} & \textbf{bowl0} & \textbf{bowl1} & \textbf{bowl2} & \textbf{bowl3} & \textbf{bowl4} & \textbf{bowl5} & \textbf{bucket0} & \textbf{bucket1} & \textbf{cap0} \\
    \midrule
    \textbf{BTF(Raw)(CVPR23’)} & 0.512  & 0.430  & 0.551  & 0.491  & \textcolor[rgb]{ 0,  .439,  .753}{\textbf{0.720 }} & 0.524  & 0.464  & 0.426  & \textcolor[rgb]{ 0,  .439,  .753}{\textbf{0.685 }} & 0.563  & 0.517  & 0.617  & 0.686  & 0.524  \\
    \textbf{BTF(FPFH)(CVPR23’)} & 0.624  & \textcolor[rgb]{ 1,  0,  0}{\textbf{0.746 }} & 0.641  & 0.549  & 0.622  & 0.710  & \textcolor[rgb]{ 1,  0,  0}{\textbf{0.768 }} & 0.518  & 0.590  & 0.679  & 0.699  & 0.401  & 0.633  & \textcolor[rgb]{ 1,  0,  0}{\textbf{0.730 }} \\
    \textbf{M3DM(CVPR23’)} & 0.577  & 0.637  & 0.663  & 0.637  & 0.532  & 0.658  & 0.663  & \textcolor[rgb]{ 0,  .439,  .753}{\textbf{0.694 }} & 0.657  & 0.624  & 0.489  & \textcolor[rgb]{ 0,  .439,  .753}{\textbf{0.698 }} & 0.699  & 0.531  \\
    \textbf{PatchCore(FPFH)(CVPR22’)} & 0.597  & 0.574  & 0.654  & 0.687  & 0.512  & 0.524  & 0.531  & 0.625  & 0.327  & 0.720  & 0.358  & 0.459  & 0.571  & 0.472  \\
    \textbf{PatchCore(PointMAE)(CVPR22’)} & 0.495  & 0.674  & 0.553  & 0.606  & 0.653  & 0.527  & 0.524  & 0.515  & 0.581  & 0.501  & 0.562  & 0.586  & 0.574  & 0.544  \\
    \textbf{CPMF(PR24’)} & 0.615  & 0.655  & 0.521  & 0.571  & 0.435  & 0.745  & 0.488  & 0.635  & 0.641  & 0.683  & 0.684  & 0.486  & 0.601  & 0.601  \\
    \textbf{Reg3D-AD(NeurIPS23’)} & \textcolor[rgb]{ 0,  .439,  .753}{\textbf{0.698 }} & 0.715  & \textcolor[rgb]{ 1,  0,  0}{\textbf{0.886 }} & 0.696  & 0.525  & \textcolor[rgb]{ 0,  .439,  .753}{\textbf{0.775 }} & 0.615  & 0.593  & 0.654  & \textcolor[rgb]{ 0,  .439,  .753}{\textbf{0.800 }} & 0.691  & 0.619  & 0.752  & 0.632  \\
    \textbf{IMRNet(CVPR24’)} & 0.671  & 0.668  & 0.556  & 0.702  & 0.641  & \textcolor[rgb]{ 1,  0,  0}{\textbf{0.781 }} & 0.705  & 0.684  & 0.599  & 0.576  & 0.715  & 0.585  & \textcolor[rgb]{ 1,  0,  0}{\textbf{0.774 }} & \textcolor[rgb]{ 0,  .439,  .753}{\textbf{0.715 }} \\
    \textbf{ISMP(AAAI25')} & \textcolor[rgb]{ 1,  0,  0}{\textbf{0.865 }} & \textcolor[rgb]{ 0,  .439,  .753}{\textbf{0.734 }} & 0.722  & \textcolor[rgb]{ 1,  0,  0}{\textbf{0.869 }} & \textcolor[rgb]{ 1,  0,  0}{\textbf{0.740 }} & 0.762  & 0.702  & \textcolor[rgb]{ 1,  0,  0}{\textbf{0.706 }} & \textcolor[rgb]{ 1,  0,  0}{\textbf{0.851 }} & 0.753  & \textcolor[rgb]{ 0,  .439,  .753}{\textbf{0.733 }} & 0.545  & 0.683  & 0.672  \\
    \textbf{DU-Net (Ours)} & 0.612  & 0.628  & \textcolor[rgb]{ 0,  .439,  .753}{\textbf{0.749 }} & \textcolor[rgb]{ 0,  .439,  .753}{\textbf{0.822 }} & 0.641  & 0.769  & \textcolor[rgb]{ 0,  .439,  .753}{\textbf{0.735 }} & 0.617  & 0.574  & \textcolor[rgb]{ 1,  0,  0}{\textbf{0.812 }} & \textcolor[rgb]{ 1,  0,  0}{\textbf{0.744 }} & \textcolor[rgb]{ 1,  0,  0}{\textbf{0.738 }} & \textcolor[rgb]{ 0,  .439,  .753}{\textbf{0.754 }} & 0.701  \\
    \midrule

    \midrule
    \textbf{Method} & \textbf{cap3} & \textbf{cap4} & \textbf{cap5} & \textbf{cup0} & \textbf{cup1} & \textbf{eraser0} & \textbf{headset0} & \textbf{headset1} & \textbf{helmet0} & \textbf{helmet1} & \textbf{helmet2} & \textbf{helmet3} & \textbf{jar0} & \textbf{micro.} \\
    \midrule
    \textbf{BTF(Raw)(CVPR23’)} & 0.687  & 0.469  & 0.373  & 0.632  & 0.561  & 0.637  & 0.578  & 0.475  & 0.504  & 0.449  & 0.605  & 0.700  & 0.423  & 0.583  \\
    \textbf{BTF(FPFH)(CVPR23’)} & 0.658  & 0.524  & 0.586  & \textcolor[rgb]{ 1,  0,  0}{\textbf{0.790 }} & 0.619  & 0.719  & 0.620  & 0.591  & 0.575  & \textcolor[rgb]{ 1,  0,  0}{\textbf{0.749 }} & 0.643  & \textcolor[rgb]{ 0,  .439,  .753}{\textbf{0.724 }} & 0.427  & 0.675  \\
    \textbf{M3DM(CVPR23’)} & 0.605  & 0.718  & 0.655  & 0.715  & 0.556  & 0.710  & 0.581  & 0.585  & 0.599  & 0.427  & 0.623  & 0.655  & 0.541  & 0.358  \\
    \textbf{PatchCore(FPFH)(CVPR22’)} & 0.653  & 0.595  & \textcolor[rgb]{ 0,  .439,  .753}{\textbf{0.795 }} & 0.655  & 0.596  & \textcolor[rgb]{ 1,  0,  0}{\textbf{0.810 }} & 0.583  & 0.464  & 0.548  & 0.489  & 0.455  & \textcolor[rgb]{ 1,  0,  0}{\textbf{0.737 }} & 0.478  & 0.488  \\
    \textbf{PatchCore(PointMAE)(CVPR22’)} & 0.488  & 0.725  & 0.545  & 0.510  & \textcolor[rgb]{ 1,  0,  0}{\textbf{0.856 }} & 0.378  & 0.575  & 0.423  & 0.580  & 0.562  & 0.651  & 0.615  & 0.487  & \textcolor[rgb]{ 1,  0,  0}{\textbf{0.886 }} \\
    \textbf{CPMF(PR24’)} & 0.551  & 0.553  & 0.551  & 0.497  & 0.509  & 0.689  & 0.699  & 0.458  & 0.555  & 0.542  & 0.515  & 0.520  & 0.611  & 0.545  \\
    \textbf{Reg3D-AD(NeurIPS23’)} & 0.718  & \textcolor[rgb]{ 1,  0,  0}{\textbf{0.815 }} & 0.467  & 0.685  & 0.698  & \textcolor[rgb]{ 0,  .439,  .753}{\textbf{0.755 }} & 0.580  & 0.626  & 0.600  & 0.624  & \textcolor[rgb]{ 1,  0,  0}{\textbf{0.825 }} & 0.620  & 0.599  & 0.599  \\
    \textbf{IMRNet(CVPR24’)} & 0.706  & 0.753  & 0.742  & 0.643  & 0.688  & 0.548  & \textcolor[rgb]{ 0,  .439,  .753}{\textbf{0.705 }} & 0.476  & 0.598  & 0.604  & 0.644  & 0.663  & \textcolor[rgb]{ 0,  .439,  .753}{\textbf{0.765 }} & \textcolor[rgb]{ 0,  .439,  .753}{\textbf{0.742 }} \\
    \textbf{ISMP(AAAI25')} & \textcolor[rgb]{ 1,  0,  0}{\textbf{0.775 }} & 0.661  & 0.770  & 0.552  & \textcolor[rgb]{ 0,  .439,  .753}{\textbf{0.851 }} & 0.524  & 0.472  & \textcolor[rgb]{ 1,  0,  0}{\textbf{0.843 }} & \textcolor[rgb]{ 0,  .439,  .753}{\textbf{0.615 }} & 0.603  & 0.568  & 0.522  & 0.661  & 0.600  \\
    \textbf{DU-Net (Ours)} & \textcolor[rgb]{ 0,  .439,  .753}{\textbf{0.763 }} & \textcolor[rgb]{ 0,  .439,  .753}{\textbf{0.783 }} & \textcolor[rgb]{ 1,  0,  0}{\textbf{0.844 }} & \textcolor[rgb]{ 0,  .439,  .753}{\textbf{0.727 }} & 0.654  & 0.569  & \textcolor[rgb]{ 1,  0,  0}{\textbf{0.718 }} & \textcolor[rgb]{ 0,  .439,  .753}{\textbf{0.749 }} & \textcolor[rgb]{ 1,  0,  0}{\textbf{0.718 }} & \textcolor[rgb]{ 0,  .439,  .753}{\textbf{0.737 }} & \textcolor[rgb]{ 0,  .439,  .753}{\textbf{0.744 }} & 0.682  & \textcolor[rgb]{ 1,  0,  0}{\textbf{0.771 }} & 0.648  \\
    \midrule

    \midrule
    \textbf{Method} & \textbf{shelf0} & \textbf{tap0} & \textbf{tap1} & \textbf{vase0} & \textbf{vase1} & \textbf{vase2} & \textbf{vase3} & \textbf{vase4} & \textbf{vase5} & \textbf{vase7} & \textbf{vase8} & \textbf{vase9} & \textbf{Average} & \textbf{Mean Rank} \\
    \midrule
    \textbf{BTF(Raw)(CVPR23’)} & 0.464  & 0.527  & 0.564  & 0.618  & 0.549  & 0.403  & 0.602  & 0.613  & 0.585  & 0.578  & 0.550  & 0.564  & 0.550  & 7.575  \\
    \textbf{BTF(FPFH)(CVPR23’)} & 0.619  & 0.568  & 0.596  & 0.642  & 0.619  & 0.646  & \textcolor[rgb]{ 0,  .439,  .753}{\textbf{0.699}} & 0.710  & 0.429  & 0.540  & 0.662  & 0.568  & 0.628  & 5.275  \\
    \textbf{M3DM(CVPR23’)} & 0.554  & 0.654  & 0.712  & 0.608  & 0.602  & 0.737  & 0.658  & 0.655  & 0.642  & 0.517  & 0.551  & 0.663  & 0.616  & 5.725  \\
    \textbf{PatchCore(FPFH)(CVPR22’)} & 0.613  & 0.733  & \textcolor[rgb]{ 1,  0,  0}{\textbf{0.768}} & 0.655  & 0.453  & 0.721  & 0.430  & 0.505  & 0.447  & 0.693  & 0.575  & 0.663  & 0.580  & 6.650  \\
    \textbf{PatchCore(PointMAE)(CVPR22’)} & 0.543  & \textcolor[rgb]{ 1,  0,  0}{\textbf{0.858}} & 0.541  & 0.677  & 0.551  & \textcolor[rgb]{ 0,  .439,  .753}{\textbf{0.742}} & 0.465  & 0.523  & 0.572  & 0.651  & 0.364  & 0.423  & 0.577  & 6.875  \\
    \textbf{CPMF(PR24’)} & \textcolor[rgb]{ 1,  0,  0}{\textbf{0.783}} & 0.458  & 0.657  & 0.458  & 0.486  & 0.582  & 0.582  & 0.514  & 0.651  & 0.504  & 0.529  & 0.545  & 0.573  & 7.325  \\
    \textbf{Reg3D-AD(NeurIPS23’)} & 0.688  & 0.589  & 0.741  & 0.548  & 0.602  & 0.405  & 0.511  & \textcolor[rgb]{ 0,  .439,  .753}{\textbf{0.755}} & 0.624  & \textcolor[rgb]{ 1,  0,  0}{\textbf{0.881}} & \textcolor[rgb]{ 1,  0,  0}{\textbf{0.811}} & \textcolor[rgb]{ 0,  .439,  .753}{\textbf{0.694}} & 0.668  & 4.125  \\
    \textbf{IMRNet(CVPR24’)} & 0.605  & 0.681  & 0.699  & 0.535  & \textcolor[rgb]{ 1,  0,  0}{\textbf{0.685}} & 0.614  & 0.401  & 0.524  & \textcolor[rgb]{ 0,  .439,  .753}{\textbf{0.682}} & 0.593  & 0.635  & 0.691  & 0.650  & 4.525  \\
    \textbf{ISMP(AAAI25')} & 0.701  & \textcolor[rgb]{ 0,  .439,  .753}{\textbf{0.844}} & 0.678  & \textcolor[rgb]{ 0,  .439,  .753}{\textbf{0.687}} & 0.534  & \textcolor[rgb]{ 1,  0,  0}{\textbf{0.773}} & 0.622  & 0.546  & 0.580  & \textcolor[rgb]{ 0,  .439,  .753}{\textbf{0.747}} & 0.736  & \textcolor[rgb]{ 1,  0,  0}{\textbf{0.823}} & \textcolor[rgb]{ 0,  .439,  .753}{\textbf{0.691 }} & \textcolor[rgb]{ 0,  .439,  .753}{\textbf{3.925}} \\
    \textbf{DU-Net (Ours)} & \textcolor[rgb]{ 0,  .439,  .753}{\textbf{0.740}} & 0.728  & \textcolor[rgb]{ 0,  .439,  .753}{\textbf{0.743}} & \textcolor[rgb]{ 1,  0,  0}{\textbf{0.699}} & \textcolor[rgb]{ 0,  .439,  .753}{\textbf{0.648}} & 0.650  & \textcolor[rgb]{ 1,  0,  0}{\textbf{0.731}} & \textcolor[rgb]{ 1,  0,  0}{\textbf{0.765}} & \textcolor[rgb]{ 1,  0,  0}{\textbf{0.711}} & 0.728  & \textcolor[rgb]{ 0,  .439,  .753}{\textbf{0.762}} & 0.581  & \textcolor[rgb]{ 1,  0,  0}{\textbf{0.712}} & \textcolor[rgb]{ 1,  0,  0}{\textbf{2.900}} \\
    \bottomrule
    \end{tabular}%
}
  \label{results2}
\end{table*}

\begin{table*}[!ht]
  \centering
  \caption{P-AUROC performance of different methods on Real3D-AD across 12 categories, where best and second-place results are highlighted in \textcolor[rgb]{1, 0, 0}{\textbf{red}} and \textcolor[rgb]{0, .439, .753}{\textbf{blue}}, respectively.}
  \resizebox{0.95\textwidth}{!}{
    \begin{tabular}{c|cccccccccccccc}
    \toprule
    \multicolumn{14}{c}{\textbf{P-AUROC}}                                                                         &  \\
    \midrule
    \textbf{Method} & \textbf{Airplane} & \textbf{Car} & \textbf{Candy} & \textbf{Chicken} & \textbf{Diamond} & \textbf{Duck} & \textbf{Fish} & \textbf{Gemstone} & \textbf{Seahorse} & \textbf{Shell} & \textbf{Starfish} & \textbf{Toffees} & \textbf{Mean} & \textbf{Mean Rank} \\
    \midrule
    \textbf{BTF(Raw) (CVPR23')} & 0.564  & 0.647  & 0.735  & 0.609  & 0.563  & 0.601  & 0.514  & 0.597  & 0.520  & 0.489  & 0.392  & 0.623  & 0.571  & 7.538  \\
    \textbf{BTF(FPFH) (CVPR23')} & \textcolor[rgb]{ 0,  .439,  .753}{\textbf{0.738}} & 0.708  & 0.864  & 0.735  & 0.882  & \textcolor[rgb]{ 0,  .439,  .753}{\textbf{0.875}} & 0.709  & \textcolor[rgb]{ 0,  .439,  .753}{\textbf{0.891}} & 0.512  & 0.571  & 0.501  & 0.815  & 0.733  & 4.385  \\
    \textbf{M3DM (CVPR23')} & 0.547  & 0.602  & 0.679  & 0.678  & 0.608  & 0.667  & 0.606  & 0.674  & 0.560  & 0.738  & 0.532  & 0.682  & 0.631  & 6.385  \\
    \textbf{PatchCore(FPFH) (CVPR22')} & 0.562  & 0.754  & 0.780  & 0.429  & 0.828  & 0.264  & 0.829  & \textcolor[rgb]{ 1,  0,  0}{\textbf{0.910}} & 0.739  & 0.739  & 0.606  & 0.747  & 0.682  & 5.000  \\
    \textbf{PatchCore(PointMAE) (CVPR22')} & 0.569  & 0.609  & 0.627  & 0.729  & 0.718  & 0.528  & 0.717  & 0.444  & 0.633  & 0.709  & 0.580  & 0.580  & 0.620  & 6.462  \\
    \textbf{Reg3D-AD (NeurIPS23')} & 0.631  & 0.718  & 0.724  & 0.676  & 0.835  & 0.503  & 0.826  & 0.545  & \textcolor[rgb]{ 1,  0,  0}{\textbf{0.817}} & 0.811  & 0.617  & 0.759  & 0.705  & 4.538  \\
    \textbf{R3D-AD (ECCV24')} & 0.594  & 0.557  & 0.593  & 0.620  & 0.555  & 0.635  & 0.573  & 0.668  & 0.562  & 0.578  & 0.608  & 0.568  & 0.592  & 7.077  \\
    \textbf{ISMP (AAAI25')} & \textcolor[rgb]{ 1,  0,  0}{\textbf{0.753}} & \textcolor[rgb]{ 0,  .439,  .753}{\textbf{0.836}} & \textcolor[rgb]{ 1,  0,  0}{\textbf{0.907}} & \textcolor[rgb]{ 0,  .439,  .753}{\textbf{0.798}} & \textcolor[rgb]{ 0,  .439,  .753}{\textbf{0.926}} & \textcolor[rgb]{ 1,  0,  0}{\textbf{0.876}} & \textcolor[rgb]{ 0,  .439,  .753}{\textbf{0.886}} & 0.857  & \textcolor[rgb]{ 0,  .439,  .753}{\textbf{0.813}} & \textcolor[rgb]{ 0,  .439,  .753}{\textbf{0.839}} & \textcolor[rgb]{ 0,  .439,  .753}{\textbf{0.641}} & \textcolor[rgb]{ 1,  0,  0}{\textbf{0.895}} & \textcolor[rgb]{ 0,  .439,  .753}{\textbf{0.836}} & \textcolor[rgb]{ 0,  .439,  .753}{\textbf{1.769}} \\
    \textbf{DU-Net (Ours)} & 0.721  & \textcolor[rgb]{ 1,  0,  0}{\textbf{0.896}} & \textcolor[rgb]{ 0,  .439,  .753}{\textbf{0.884}} & \textcolor[rgb]{ 1,  0,  0}{\textbf{0.838}} & \textcolor[rgb]{ 1,  0,  0}{\textbf{0.938}} & 0.793  & \textcolor[rgb]{ 1,  0,  0}{\textbf{0.910}} & 0.848  & 0.801  & \textcolor[rgb]{ 1,  0,  0}{\textbf{0.872}} & \textcolor[rgb]{ 1,  0,  0}{\textbf{0.799}} & \textcolor[rgb]{ 0,  .439,  .753}{\textbf{0.861}} & \textcolor[rgb]{ 1,  0,  0}{\textbf{0.847}} & \textcolor[rgb]{ 1,  0,  0}{\textbf{1.846}} \\
    \bottomrule
    \end{tabular}%
}
  \label{results11}
\end{table*}
\section{Experiments}
\subsection{Implementation}
\textbf{Datasets.}
We conducted extensive experiments on Anomaly-Shape Net~\cite{IMRNet} and Real3D-AD~\cite{Real3D-AD}. (1) The Anomaly-ShapeNet has over 1,600 positive and negative samples from 40 categories, leading to a more challenging setting due to the large number of categories. (2) The Real3D-AD consists of 1,254 large-scale high-resolution point cloud samples from 12 categories, with the training set for each category only containing 4 normal samples.

\textbf{Baselines.} We selected BTF~\cite{BTF}, PatchCore~\cite{Patchcore}, M3DM~\cite{M3DM}, CPMF~\cite{CPMF}, Reg 3D-AD~\cite{Real3D-AD}, IMRNet~\cite{IMRNet}, R3D-AD~\cite{R3DAD}, and ISMP~\cite{ISMP} for comparison. The results of these methods were obtained by executing publicly available code or by referring to their papers.

\textbf{Evaluation Metrics.} We adopted Object-Level Area Under the Receiver Operator Curve~(O-AUROC, $\uparrow$) and Object-Level Area Under the Per-Region-Overlap~(O-AUPR, $\uparrow$) to evaluate object-level anomaly detection performance, and Point-Level Area Under the Receiver Operator Curve~(P-AUROC, $\uparrow$) and Point-Level Area Under the Per-Region-Overlap~(P-AUPR, $\uparrow$) to evaluate pixel-level anomaly segmentation precision. To avoid a small number of categories dominating the average performance, the average ranking ($\downarrow$) was used for evaluation. In addition, Frames Per Second~(FPS, $\uparrow$) was introduced to compare the inference speed of different methods.

\textbf{Experimental Details.} Experiments were carried out on a machine equipped with two L20~(48GB) GPUs to promote stable training with a larger batch size and evaluated on only an RTX3090~(24GB) GPU. We pre-trained the proposed Down-Net on the ShapeNetPart dataset~\cite{shapenet} like PointMAE~\cite{pointmae} and trained Up-Net on a subset of the Visionair repository following the setup of PU-Net~\cite{yu2018pu}. The parameters $\alpha$ and $\beta$ in Noisy-Gen were set to 0.08 and 0.15, respectively. The number of groups $G$ and the number of points for each group $K$ were set to 8192 and 640 for Down-Net, and the up-sampling rate $\gamma$ was set to 8 for Up-Net. Three independent runs of experiments were conducted, and their performance was averaged to obtain convincing and representative results.

\subsection{Main Results}
\textbf{Results on Anomaly-ShapeNet.}
Tables~\ref{results1} and \ref{results2} present the detection and segmentation results of different methods on Anomaly-ShapeNet, respectively. Our method achieved an average O-AUROC and P-AUROC of 0.797 and 0.712, outperforming the second-best methods by 2.1\% and 3.8\%, respectively. Notably, our method obtained the best average ranking on both O-AUROC and P-AUROC, demonstrating the superior generalization on different categories. 
 
\textbf{Results on Real3D-AD.} Tables~\ref{results3} and ~\ref{results11} report the performance of different methods on Real3D-AD dataset. Our method obtained an impressive O-AUROC and P-AUROC of 0.795 and 0.847, respectively, surpassing the best baseline by 3.8\% and 0.9\%, respectively. Moreover, our approach yielded the best average rankings on both metrics, indicating its effectiveness for high-resolution point clouds.

Existing methods face challenges in precisely reconstructing the point cloud due to the significant difference between the training and test data. Our method uses the Down-Net to keep the group centers for main structure consistency and the Up-Net to reconstruct high-precision point clouds by fusing multi-scale features, thereby leading to better detection and segmentation performance. 

\subsection{Ablation Studies}

Our method consists of three key components: the \textbf{Noise-Gen}, the \textbf{Down-Net} with the losses $\mathcal{L}_{MSE}$, $\mathcal{L}_{COS}$, and $\mathcal{L}_{CD}$, and the \textbf{Up-Net} with the losses $\mathcal{L}_{REP}$ and $\mathcal{L}_{EMD}$. To evaluate the importance of these components, we conducted ablation studies on Anomaly-ShapeNet across 40 classes, and the results are shown in Table \ref{results4}, where $M_{i}$ denotes the method without the corresponding loss, $M_{w/o}$ represents the method without the corresponding component, and $M_{replace \ D}$ stands for the method by replacing the Down-Net module with FPS down-sampling.

\textbf{The losses for Down-Net.} 
For Down-Net, training without the reconstruction distance loss $\mathcal{L}_{MSE}$ ($M_1$) resulted in a performance degradation of 11.7\%, 7.0\%, 12.0\%, and 15.2\% in O-AUROC, P-AUROC, O-AUPR, and P-AUPR, respectively. Removing the orientation constraint loss $\mathcal{L}_{COS}$ ($M_2$) led to a performance reduction of 3.1\%, 1.0\%, 5.8\%, and 6.2\% in four metrics, respectively. While the Down-Net without the loss $\mathcal{L}_{CD}$ ($M_3$) also became worse, causing a decrease of 3.4\%, 3.1\%, 6.0\%, and 9.7\% in different metrics, respectively. These results clearly demonstrate the significance of different losses for Down-Net in accurately preserving anomaly-free center point clouds.
 \begin{table}[t!]
  \centering
  \caption{Results of ablation experiments.}
  \resizebox{0.95\columnwidth}{!}{
    \begin{tabular}{c|ccccc|cccc}
    \toprule
    \textbf{Method} & $M_1$ & $M_2$ & $M_3$ & $M_4$ & $M_5$ & $M_{w/o \ N}$ & $M_{Replace \ D}$ & $M_{w/o \ U}$ &\cellcolor{gray!20}Ours\\
    \midrule
    $\mathcal{L}_{MSE}$    & \ensuremath{\times} & \checkmark & \checkmark  & \checkmark  & \checkmark & \checkmark & \ensuremath{\times} & \checkmark&  \cellcolor{gray!20}\checkmark\\
    $\mathcal{L}_{COS}$   & \checkmark & \ensuremath{\times} & \checkmark & \checkmark & \checkmark & \checkmark & \ensuremath{\times}& \checkmark& \cellcolor{gray!20}\checkmark\\
    $\mathcal{L}_{CD}$ & \checkmark & \checkmark &\ensuremath{\times}& \checkmark  & \checkmark & \checkmark& \ensuremath{\times}& \checkmark& \cellcolor{gray!20}\checkmark \\
    \midrule
    $\mathcal{L}_{REP}$   & \checkmark  & \checkmark & \checkmark & \ensuremath{\times} & \checkmark & \checkmark& \checkmark& \ensuremath{\times}& \cellcolor{gray!20}\checkmark \\
    $\mathcal{L}_{EMD}$   & \checkmark & \checkmark & \checkmark & \checkmark & \ensuremath{\times} & \checkmark& \checkmark& \ensuremath{\times}& \cellcolor{gray!20}\checkmark \\
    \midrule
    Noise-Gen   & \checkmark & \checkmark & \checkmark  & \checkmark  & \checkmark & \ensuremath{\times} & \checkmark & \checkmark&  \cellcolor{gray!20}\checkmark\\
    \midrule
    O-AUROC & 0.680 &0.766&0.763& 0.733 &0.573 &0.469& 0.550 & 0.620   & \cellcolor{gray!20}\textbf{0.797}\\
    P-AUROC & 0.642 &0.702&0.681& 0.650 &0.543&0.511& 0.524 & 0.588   & \cellcolor{gray!20}\textbf{0.712}\\
    O-AUPR &  0.682 &0.742&0.744& 0.702 &0.596 &0.482& 0.534 &  0.613   &\cellcolor{gray!20}\textbf{0.802}\\
    P-AUPR & 0.092 &0.182& 0.147& 0.104 &0.048&0.013& 0.028 &  0.080   &\cellcolor{gray!20}\textbf{0.244}\\
    \bottomrule
    \end{tabular}%
 }
  \label{results4}
\end{table}

\textbf{The losses for Up-Net.}
For Up-Net, training without the repulsion constraints loss $\mathcal{L}_{REP}$ ($M_4$) caused a performance decrease of 6.4\%, 6.2\%, 10.0\%, and 14.0\% in O-AUROC, P-AUROC, O-AUPR, and P-AUPR, respectively. The Up-Net without the overall reconstruction constraints loss $\mathcal{L}_{EMD}$ ($M_5$) results in a performance reduction of 22.4\%, 16.9\%, 20.6\%, and 19.6\% in four metrics, respectively. These results indicate the importance of different losses for Up-Net to reconstruct high-precision point cloud with multi-scale features.

\textbf{The effect of each module.} 
For the components of our method, the Noise-Gen module is used to inject noise on normal patches, and the performance of our method without Noise-Gen $M_{w/o\ N}$ was dramatically declined by 32.8\%, 20.1\%, 32.0\%, and 23.1\% in O-AUROC, P-AUROC, O-AUPR, and P-AUPR, respectively. Replacing the Down-Net module with FPS down-sampling $M_{replace \ D}$ caused a performance degradation in four metrics by 24.7\%, 18.8\%, 26.8\%, and 21.6\%, respectively. While removing the Up-Net from our method $M_{w/o\ U}$ also deteriorated performance, decreasing by 17.7\%, 12.4\%, 18.9\%, and 16.4\%, respectively. These results highlight that both components contribute to detection performance, with their combination providing the most significant performance improvements. More ablation results are reported in the \textit{Supplementary Material}.


\subsection{Parameter Sensitivity Analysis}
To examine the effects of different parameters on the proposed modules, we conducted parameter sensitivity experiments on Anomaly-ShapeNet.
 
\textbf{Parameter effects on Noisy-Gen.}
There are two noise parameters $\alpha$ and $\beta$ in Noisy-Gen. As shown in Table~\ref{results5}, Selecting appropriate noise parameters $\alpha$ and $\beta$ for the center and the whole Patch is beneficial for anomaly detection. When these parameters were set to $\alpha=0.08$ and $\beta=0.15$, the best performance was achieved. This may be attributed to that the moderate noise balances data diversity and reconstruction difficulty in generating noisy patches.
In addition, when replacing Gaussian noise with uniform or salt-and-pepper noises, the performance of our method was reduced on average by 19.0\%, 17.4\%, 19.5\%, and 16.9\% in O-AUROC, P-AUROC, O-AUPR, and P-AUPR, respectively. This may be attributed to that Gaussian noise is more appropriate for diversifying data in the real world and also facilitates the model to learn a robust representation. 

\begin{table}[t!]
  \centering
  \caption{Experimental results of the effect of parameters on Noisy-Gen.}
  \resizebox{0.95\columnwidth}{!}{
    \begin{tabular}{c|rrrrrrr}
    \toprule
    \textbf{Parm.}~($\alpha/\beta$) & \multicolumn{1}{c}{(0.08/0.09)} & \multicolumn{1}{c}{(0.08/0.12)} & \multicolumn{1}{c}{(0.08/0.18)} & \multicolumn{1}{c}{(0.04/0.15)} & \multicolumn{1}{c}{(0.06/0.15)} & \multicolumn{1}{c}{(0.10/0.15)} \\
    \midrule
    O-AUROC&\multicolumn{1}{c}{0.767}&\multicolumn{1}{c}{0.796}&\multicolumn{1}{c}{0.784}&\multicolumn{1}{c}{0.791}&\multicolumn{1}{c}{0.788}&\multicolumn{1}{c}{0.793}\\
    P-AUROC &\multicolumn{1}{c}{0.710}&\multicolumn{1}{c}{0.707}&\multicolumn{1}{c}{0.703}&\multicolumn{1}{c}{0.711}&\multicolumn{1}{c}{0.708}&\multicolumn{1}{c}{0.710}\\
    O-AUPR &\multicolumn{1}{c}{0.758}&\multicolumn{1}{c}{0.785}&\multicolumn{1}{c}{0.795}&\multicolumn{1}{c}{0.796}&\multicolumn{1}{c}{0.799}&\multicolumn{1}{c}{0.787}\\
    P-AUPR &\multicolumn{1}{c}{0.218}&\multicolumn{1}{c}{0.232}&\multicolumn{1}{c}{0.231}&\multicolumn{1}{c}{0.237}&\multicolumn{1}{c}{0.240}&\multicolumn{1}{c}{0.238}\\
    \midrule
    \midrule
    \textbf{Module} & \multicolumn{2}{c}{Uniform Noise} & \multicolumn{2}{c}{Salt-and-Pepper Noise} & \multicolumn{2}{c}{\cellcolor{gray!20}Ours (0.08, 0.15)} \\
    \midrule
    O-AUROC & \multicolumn{2}{c}{0.572} & \multicolumn{2}{c}{0.642} & \multicolumn{2}{c}{\cellcolor{gray!20}\textbf{0.797}} \\
    P-AURC  & \multicolumn{2}{c}{0.708} & \multicolumn{2}{c}{0.542} & \multicolumn{2}{c}{\cellcolor{gray!20}\textbf{0.712}} \\
    O-AUPR & \multicolumn{2}{c}{0.588} & \multicolumn{2}{c}{0.627} & \multicolumn{2}{c}{\cellcolor{gray!20}\textbf{0.802}} \\
    P-AUPR & \multicolumn{2}{c}{0.104} & \multicolumn{2}{c}{0.047} & \multicolumn{2}{c}{\cellcolor{gray!20}\textbf{0.244}} \\
    \bottomrule
    \end{tabular}%
 }
  \label{results5}
\end{table}  

\begin{table}[!th]
  \centering
  \caption{Experimental results of the effect of parameters on Down-Net.}
  \resizebox{0.9\columnwidth}{!}{
    \begin{tabular}{c|ccccc}
    \toprule
    \textbf{Parm.}(\textit{G}/\textit{K}) & (1024/640) & (4096/640) & \cellcolor{gray!20}\textbf{(8192/640)} & (8192/384) & (8192/128) \\
    \midrule
    O-AUROC & 0.618 & 0.707 & \cellcolor{gray!20}\textbf{0.797} & 0.724 & 0.685 \\
    P-AUROC & 0.584 & 0.638 & \cellcolor{gray!20}\textbf{0.712} & 0.69  & 0.659 \\
    O-AUPR & 0.63  & 0.681 & \cellcolor{gray!20}\textbf{0.802} & 0.708 & 0.671 \\
    P-AUPR & 0.029 & 0.133 & \cellcolor{gray!20}\textbf{0.244} & 0.208 & 0.134 \\
    FPS   & 3.4   & 3     & \cellcolor{gray!20}\textbf{2.1}   & 2.5   & 2.9 \\
    \bottomrule
    \end{tabular}%
 }
  \label{results6}
\end{table}  

\begin{table}[!ht]
  \centering
  \caption{Experimental results of the effect of parameters on Up-Net.}
  \resizebox{0.6\columnwidth}{!}{
  \begin{tabular}{c|rrrrrr}
        \toprule
        \textbf{Ratio} & $\gamma=1$ & $\gamma=2$ & $\gamma=4$ & $\gamma=6$ &\cellcolor{gray!20} $\gamma=8$ \\
        \midrule
        O-AUROC&\multicolumn{1}{c}{0.620}&\multicolumn{1}{c}{0.644}&\multicolumn{1}{c}{0.727}&\multicolumn{1}{c}{0.743}&  \multicolumn{1}{c}{\cellcolor{gray!20}\textbf{0.797}}\\
        P-AUROC&\multicolumn{1}{c}{0.588}&\multicolumn{1}{c}{0.601}&\multicolumn{1}{c}{0.637}&\multicolumn{1}{c}{0.685}&  \multicolumn{1}{c}{\cellcolor{gray!20}\textbf{0.712}}\\
        O-AUPR&\multicolumn{1}{c}{0.613}&\multicolumn{1}{c}{0.647}&\multicolumn{1}{c}{0.723}&\multicolumn{1}{c}{0.758}&\multicolumn{1}{c}{\cellcolor{gray!20}\textbf{0.802}}\\
        P-AUPR&\multicolumn{1}{c}{0.080}&\multicolumn{1}{c}{0.114}&\multicolumn{1}{c}{0.138}&\multicolumn{1}{c}{0.188}&\multicolumn{1}{c}{\cellcolor{gray!20}\textbf{0.244}}\\
        FPS&\multicolumn{1}{c}{4.4}&\multicolumn{1}{c}{3.1}&\multicolumn{1}{c}{2.6}&\multicolumn{1}{c}{2.3}&\multicolumn{1}{c}{\cellcolor{gray!20}\textbf{2.1}}\\
        \bottomrule
      \end{tabular}%
 }
  \label{results7}
\end{table} 

\textbf{Parameter effects on Down-Net.} 
The Down-Net module involves two adjustable parameters: the number of groups $G$ and the number of points $K$ in each group. As shown in Table~\ref{results6}, the increase of $G$ from 1024 to 8192 brought a performance improvement of 17.9\%, 12.8\%, 17.2\%, and 21.5\% in O-AUROC, P-AUROC, O-AUPR, and P-AUPR, respectively. Similarly, increasing points $K$ for center generation from 128 to 640 enhanced the anomaly detection metrics by 11.2\%, 7.4\%, 12.1\%, and 11.1\%, respectively. To balance computational efficiency and detection accuracy, the settings $G=8192$ and $K=512$ were adopted in the experiments.

\textbf{Parameter effects on Up-Net.}
Point cloud reconstruction in Up-Net relies on the up-sampling rate $\gamma$. As seen in Table~\ref{results7}, increasing $\gamma$ from 2 to 8 improved detection performance by 17.7\%, 12.4\%, 18.9\%, and 23.6\% in four metrics, with the inference efficiency FPS decreasing from 4.4 to 2.1. The reason for this may be that a larger up-sampling rate provides more points for the reconstruction of the high-resolution point cloud, allowing for better anomaly detection performance. But a high up-sampling rate leads to more computational costs. Thus, this parameter was limited to 8 in the experiments.

\section{Conclusion}

High-resolution point clouds are characterized by large scale and high complexity. This paper proposes a Down-Up sampling Network (DUS-Net) to address the challenge of reconstructing high-resolution point clouds for 3D anomaly detection. To diversify the training data and provide denoising ability, we introduce a Noise Generation (Noise-Gen) module to generate noisy patches and present a Down-sampling Network (Down-Net) to predict robust and anomaly-free group center points from these corrupted patches. Upon these precise center points, we design an Upsampling Network (Up-Net) to achieve high-fidelity anomaly-free point cloud reconstruction through hierarchical aggregation of multi-scale features. Extensive experiments demonstrate that the proposed method achieves SOTA performance across different evaluation metrics while maintaining strong robustness, offering a promising solution for high-resolution industrial 3D anomaly detection. \textbf{Limitation.} Two-stage reconstruction requires additional training and cost. Exploring single-stage frameworks is worthy of further research.
\section*{Acknowledgments}
This work was supported in part by the National Natural Science Foundation of China (Grant Nos. 62476171 and 62206122), Guangdong Basic and Applied Basic Research Foundation (Grant No. 2024A1515011367), National Undergraduate Training Program for Innovation and Entrepreneurship (Grant No. S202510590118), Guangdong Provincial Key Laboratory (Grant No. 2023B1212060076), Tencent ``Rhinoceros Birds” - Scientific Research Foundation for Young Teachers of Shenzhen University, and the Shenzhen Institute of Artificial Intelligence and Robotics for Society.
\bibliographystyle{ieeetr}
\balance
\bibliography{ref}

\end{document}